\documentclass[10pt,twocolumn,letterpaper]{article}

\usepackage[english]{babel}
\usepackage[utf8x]{inputenc}
\usepackage[T1]{fontenc}
\usepackage{enumitem}
\usepackage{amsmath}

\usepackage[a4paper,top=3cm,bottom=2cm,left=1.75cm,right=1.75cm,marginparwidth=1.75cm]{geometry}
\usepackage{mathtools,xparse}
\usepackage{adjustbox}

\DeclarePairedDelimiter{\norm}{\lVert}{\rVert}
\NewDocumentCommand{\normL}{ s O{} m }{%
  \IfBooleanTF{#1}{\norm*{#3}}{\norm[#2]{#3}}_{L_2(\Omega)}%
}
\DeclareMathOperator*{\argmax}{\arg\!\max}
\DeclareMathOperator*{\argmin}{\arg\!\min}
\DeclareDocumentCommand{\norm}{ O{} m }{\left\lVert#2\right\rVert_{#1}}

\usepackage{amsmath}
\usepackage{graphicx}
\usepackage[colorinlistoftodos]{todonotes}
\usepackage[colorlinks=true, allcolors=blue]{hyperref}
\usepackage{enumitem,amsthm,amssymb,bbm,dsfont,algorithm,euscript,courier,listings,xcolor,subcaption, float,blindtext,xparse,pgfplots,mathtools,algpseudocode}
\usepackage[font=scriptsize]{subcaption}

\makeatletter
\def\BState{\State\hskip-\ALG@thistlm}
\makeatother
\usepackage{pgfplots}
\pgfplotsset{compat=1.7}
\begin{document}
\title{Finding Solutions to Generative Adversarial Privacy}

\author{
	Dae Hyun Kim\thanks{Department of Computer Science, Stanford University} \\ dhkim16@stanford.edu
	\and 
	Taeyoung Kong\thanks{Department of Electrical     Engineering, Stanford University} \\ kongty@stanford.edu
    \and 
	Seungbin Jeong\footnotemark[2] \\ sjeong91@stanford.edu
}
\date{}

\maketitle

\section*{Abstract}
We present heuristics for solving the maximin problem induced by the generative adversarial privacy setting for linear and convolutional neural network (CNN) adversaries.
In the linear adversary setting, we present a greedy algorithm for approximating the optimal solution for the privatizer, which performs better as the number of instances increases.
We also provide an analysis of the algorithm to show that it not only removes the features most correlated with the private label first, but also preserves the prediction accuracy of public labels that are sufficiently independent of the features that are relevant to the private label.
In the CNN adversary setting, we present a method of hiding selected information from the adversary while preserving the others through alternately optimizing the goals of the privatizer and the adversary using neural network backpropagation.
We experimentally show that our method succeeds on a fixed adversary.

\section{Introduction}

Assume that an entity, whom we call the \textit{privatizer} holds some data it wishes to publish to the public, but that the data includes some sensitive information, such as gender or race, that the entity wants to hide from the public.
We will assume that there exists another entity, whom we call the \textit{adversary}, whose goal is to recover the sensitive information from the data published by the privatizer, and publicly available data that relates the features of the privatizer's data with the sensitive labels.

If the privatizer naively removes the sensitive labels and publishes the rest of the data as-is, the adversary can accurately recover the sensitive labels using various machine learning models with the publicly available data.
Thus, the privatizer needs to encrypt its data using a privatization function before it publishes it to the public.
We will assume the worst case for the privatizer, in which the privatization function is known to the adversary, perhaps through publications or information leakage.
The natural methodology that the adversary would use is to first simulate the privatization function on the publicly available data to obtain a relationship between the privatized data and the sensitive information.
Then he would use this relationship to predict the sensitive labels from the privatized data published by the privatizer.
The privatizer's goal is to hinder such attempt by the adversary.

This problem is a zero-sum game, in which the adversary attempts to minimize his loss on predicting the sensitive labels, while the privatizer attempts to maximize this value.
From the privatizer's perspective, finding the optimum requires solving a maximin problem: 
\[
	\argmax_P {\min_{f \in \mathcal{H}_{adv}} \mathcal{L}(f(P(X)), y)},
\]
where $P$ stands for the privatizer's function and $L$ stands for the loss function.
In addition, the privatizer wants to release useful data to the public, which we will model by restricting the amount of change in the data:
\[
	\norm[2]{X - P \left( X \right)} ^2 \leq D,
\]
for some constant $D$.
In this paper, we investigate this problem when $\mathcal{H}_{adv}$, the hypothesis class of the adversary, is either linear or convolutional neural network (CNN).

\section{Related Work}

\subsection{Minimax Optimization}
Zero-sum games are games in which the two players' utilities sum to zero, which is commonly seen in real life.
The two players in a zero-sum game are subject to solving a minimax game, in which each of the players attempt to minimize the maximum utility gained by the opponent.
Due to its omnipresence, much work has been done to accurately and efficiently solve the minimax games under variant restrictions~\cite{char1974minimax, char1978minimax}.

The setup that we are working with is a zero-sum game between the privatizer and the adversary, and this naturally gives rise to a maximin game, which is just a negation of a minimax game.

\subsection{Generative Adversarial Networks}
Recently, \textit{generative adversarial network} (GAN)~\cite{goodfellow2014explaining, szegedy2013intriguing} was proposed to generate data to simulate real data. 
The GAN solved the minimax problem between the \textit{generator} and the \textit{discriminator} to train a good generative network. 
The idea of training a generator with another network inspired us on how to train the privatizer network.
This generator tries to deceive the adversary, while the adversary tries to avoid being fooled by it. 

\subsection{Generative Adversarial Privacy}
Recent works~\cite{huang2017context, hamm2016minimax} have introduced the idea of \textit{generative adversarial privacy} (GAP) and showed that it is possible to add noise that selectively disturbs algorithms that can learn the \textit{private labels}, while preserving the utility of algorithms that can learn \textit{public labels}. 
There is also some promising results in the case in which the models used by the adversary and the ally are both linear~\cite{Xu2017cleaning}.

Our work differs from these previous works in that we aim to preserve the overall data, instead of choosing a label to preserve the accuracy for.

\section{Methods}

\subsection{Linear Adversary Case}
We will solve the privatizer's problem with lossy compression using a compression matrix $A$.
In this case, the privatizer's problem simplifies to:
\begin{align*}
	\text{Minimize} \quad & \norm[2]{y^T \left( X A A^+ \right) \left( X A A^+ \right)^+}^2, \\
    \text{subject to} \quad & \norm[Frob]{X - X A A^+ }^{2} \leq D
\end{align*}
where the compression matrix $A$ is the variable.
This problem is difficult because the variable is encapsulated in a pseudoinverse that is difficult to simplify.\footnote{One of the steps leading to the convexity results in the milestone had an error.}

We therefore look at a smaller subset of the problem by restricting the data compression with respect to the features.
Then, the problem reduces to partitioning the set of features into two sets, $R$, the set of features that are removed by the compression, and $S$, the set of features that are preserved by the compression.
The optimum of this subproblem can be found by solving
\begin{align*}
	\text{Maximize} \quad & \norm[2]{y - X_{i \in R} X_{i \in R}^+},  \\
    \text{subject to} \quad & \norm[2]{X_{i \in R}}^{2} \le D
\end{align*}
where the set of excluded features, $R$, is the variable.\footnote{If $y=X\theta$, the target function reduces further to $\norm[2]{\left( X_{i \in R} - X_{i \in S} X_{i \in S}^+ X_{i \in R} \right) \theta_{i \in R}}^2$.}

Solving this problem by brute force results in exponential time complexity.
Therefore, we propose an approximation algorithm for computing the optimum based on the ideas of the greedy algorithm.
In each step of the greedy algorithm (Algorithm~\ref{alg:greedy}), we select the element that maximizes the utility, $\|{y - X_{i \in R \cup \left\{ e \right\} }X_{i \in R \cup \left\{ e \right\} }^+ } \|_{2}$, per unit cost, $\norm[2]{X_{R \cup \left\{ e \right\} } }^{2}$.

\begin{algorithm}
\caption{Greedy Algorithm}\label{alg:greedy}
  \begin{algorithmic}[1]
  \Procedure{Greedy-Approx($D$)}{}
  \State $R \gets \phi $
  \State \textbf{do}
  \State \quad\, $e_{next} \gets \textsc{Find-Next}\left(R, D \right)$
  \State \quad\, $R \gets R \cup \left\{ e_{next} \right\}$
  \State \textbf{while} $e_{next} \ne \text{null}$
  \State \Return $R$
  \EndProcedure
  \end{algorithmic} 
  \begin{algorithmic}[1]
  \Procedure{Find-Next($R$, $D$)}{}
  \State ${(u/c)}_{max} \gets -\infty$
  \State $e_{next} \gets \text{null}$
  \For{$e \in \left\{1, \cdots , n\right\} \backslash R$}
  	\State $c \gets \norm[2]{X_{R \cup \left\{ e \right\} } }^{2}$
    \If{$c > D$}
    	\State \textbf{continue}
    \EndIf
    \State $u \gets \|{y - X_{i \in R \cup \left\{ e \right\} }X_{i \in R \cup \left\{ e \right\} }^+ } \|_{2}$
    \If{$v/c > (u/c)_{max}$}
    	\State $(u/c)_{max} \gets u/c$
        \State $e_{next} = e$
    \EndIf
  \EndFor
  \State \Return $e_{next}$
  \EndProcedure
  \end{algorithmic}
\end{algorithm}

\subsection{CNN Adversary Case}

\begin{figure}[ht]
	\begin{subfigure}{.5\textwidth}
    \centering
	\includegraphics[width = \linewidth]{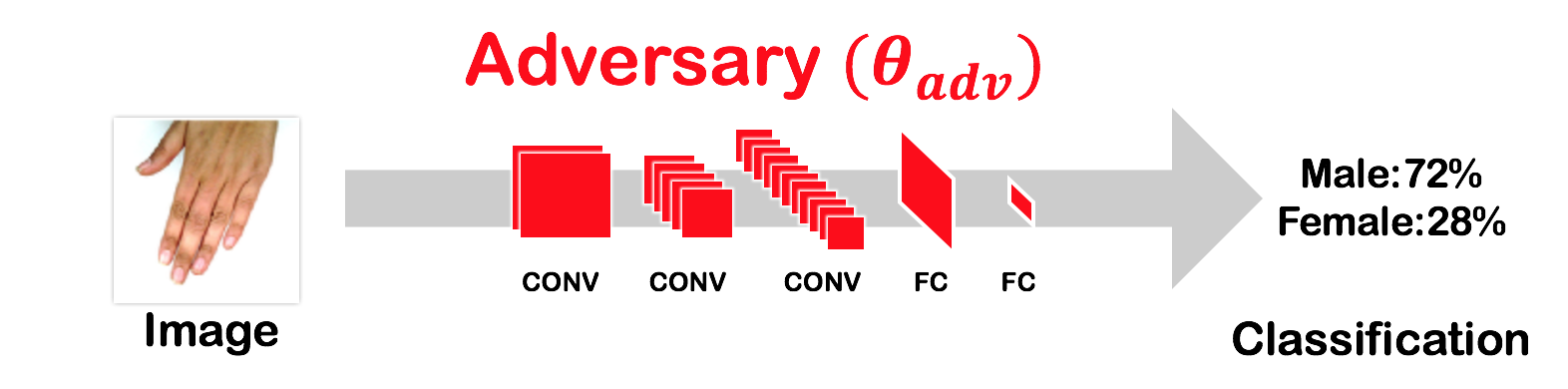}
	\caption{Adversary neural network structure}
	\label{fig:adversary}
    \end{subfigure}
	\begin{subfigure}{.5\textwidth}
	\centering
	\includegraphics[width = \linewidth]{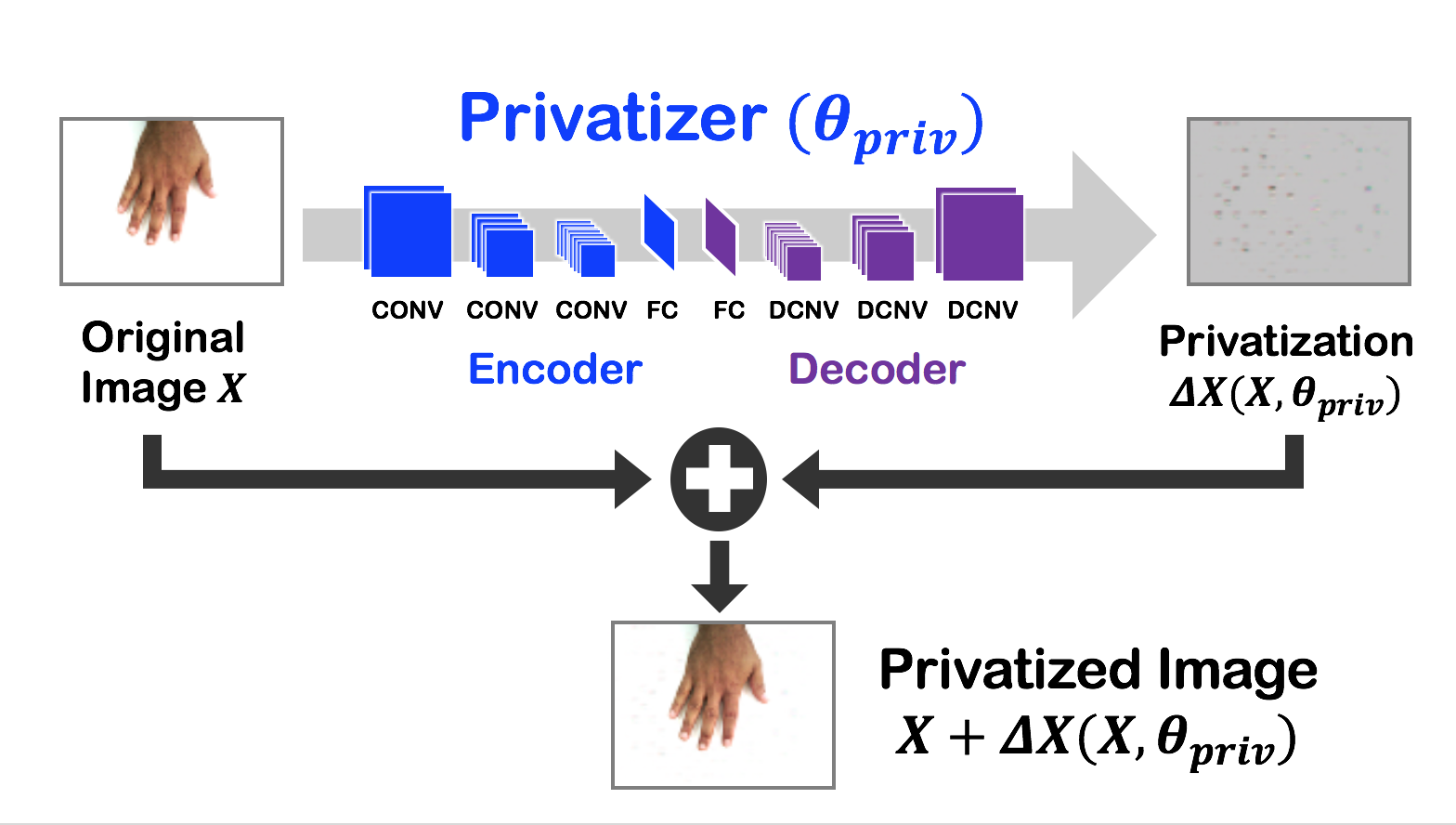}
	\caption{Privatizer neural network structure}
	\label{fig:privatizer}
	\end{subfigure}
    \caption{Adversary and privatizer network structures}
    \label{fig:cnn_struct}
\end{figure}

In the CNN adversary case, we assume that the adversary's network comprises of convolutional layers, activation layers (ReLu), pooling layers, fully connected layers, and a loss layer (softmax cross entropy loss) as exemplified by Figure \ref{fig:adversary}.



The privatizer uses a neural network (Figure \ref{fig:privatizer}) that consists of an (1) encoder, which not only reduces the features by convolutional layers, but also tries to capture the high level features that the adversary would capture, a (2) decoder, which uses deconvolutional layers to map the captured features into the pixel space.
The resulting output of the privatizer's network is the noise, $\Delta X$ that he/she adds to the original image $X$, to obtain a privatized image, $X + \Delta X$.
The restriction in the amount of change in the data, in this case, becomes:
\[
	\norm[2]{\Delta X} ^2 \leq D.
\]
We also added a constraint that we want to preserve the prediction accuracy as much as possible for a selected set of labels with a pre-trained model on the original data, which we call \textit{protected labels}.~\footnote{We are able to preserve prediction accuracies of labels that are not highly correlated with the private label without this constraint, but with this constraint, we can preserve prediction accuracies on even highly correlated labels such as gender and decoration in 11k Hands.}



\begin{algorithm}
\caption{Maximin Algorithm for CNN Adversary}\label{alg:maximin}
  \begin{algorithmic}[1]
  \Procedure{Solve-Maximin(X, D)}{}
  \State $\theta_{pro} \gets \displaystyle \argmin_{\theta} \mathcal{L}_{pro}(X; \theta)$
  \State $\theta_{adv} \sim \text{Xavier}$
  \State $\theta_{priv} \sim \text{Xavier}$
  \While{ $\Delta \theta_{adv} > \epsilon \text{ or } \Delta \theta_{priv} > \epsilon$}
  \State $\theta_{adv} \gets \displaystyle \argmin_{\theta}{\mathcal{L}_{priv} \left( X + \Delta X \left(X, \theta_{priv}, D \right); \theta \right)}$
  \State $\theta_{priv} \gets \displaystyle \argmax_{\theta}{\mathcal{L}_{priv} \left( X + \Delta X \left(X, \theta, D \right); \theta_{adv} \right)}$
  \State $\theta_{priv} \gets \displaystyle \argmin_{\theta}{\mathcal{L}_{pro} \left( X + \Delta X \left(X, \theta, D \right); \theta_{pro} \right)}$
  \EndWhile
  \EndProcedure
  \end{algorithmic}
\end{algorithm}

In determining the model for the privatization function, the privatizer solves the maximin problem mentioned in Section 1.
The privatizer's method for solving this maximin problem by (Algorithm~\ref{alg:maximin}):
(1) initializing the privatizer's internal model for the adversary's parameters, $\theta_{adv}$, and the privatizer's own parameters, $\theta_{priv}$, randomly,
(2) fixing $\theta_{priv}$ and solves for the optimal $\theta_{adv}$ against the loss function of the private labels from the adversary's perspective using backpropagation,
(3) fixing $\theta_{adv}$ and solves for the optimal $\theta_{priv}$ against the loss function of the private labels from the privatizer's perspective using backpropagation,
(4) finding the optimal $\theta_{priv}$ against the loss function of the protected labels,
and (5) repeating these steps until convergence.


\section{Dataset and Features}

\begin{figure}[t!]
\begin{center}
  \begin{adjustbox}{max width=3.2in}
  \begin{tabular}{ | c | c | c | c |}
  \hline
  dataset & instances ($m$) & features ($n$) & labels ($N$) \\ \hline \hline
  \multicolumn{4}{|c|}{$\mathcal{H}_{adv}$: linear} \\ \hline
  Beijing PM 2.5~\cite{Liang2015beijing} & $41757$ & $7$ & $2$ \\ \hline
  UCI Wine Quality~\cite{cortez2009wine} & $4898$ & $7$ & $5$ \\ \hline
  \multicolumn{4}{|c|}{$\mathcal{H}_{adv}$: CNN} \\ \hline
  GENKI~\cite{GENKI-4K} & $4000$ & $200 \times 200 \times 1$ & $4$ \\ \hline
  11k Hands~\cite{afifi2016hands} & $5538$ & $200 \times 200 \times 3$ & $4$ \\
  \hline
  \end{tabular}
  \end{adjustbox}
  \caption{\label{fig:dataset}Summary of datasets}
\end{center}
\end{figure}

For the linear adversary case, we use generated data of various sizes to confirm that the greedy approximation indeed approaches the true optimum.
We generated the input data randomly using a uniform distribution $\text{unif} \left( 0, 1 \right)$, and varied the number of features from $4$ to $7$ and the number of instances from $10$ to $1000$.

We use Beijing PM 2.5 dataset~\cite{Liang2015beijing} and UCI Wine Quality dataset~\cite{cortez2009wine} (Figure~\ref{fig:dataset}) to observe the properties of the optimum that we obtain.
Of the properties included in the Beijing PM 2.5 dataset, we utilize the properties \textit{year}, \textit{month}, \textit{day}, \textit{hour},  \textit{temperature}, \textit{pressure}, and \textit{cumulated wind speed} as features, and \textit{PM2.5 concentration} and \textit{dew point} as labels.
For this dataset, we removed any instance that included missing features.
In addition, of the properties included in the UCI Wine Quality dataset, we utilize the properties \textit{fixed acidity}, \textit{volatile acidity}, \textit{citric acid}, \textit{residual sugar}, \textit{chlorides}, \textit{free sulfur dioxide}, and \textit{total sulfur dioxide} as features, and \textit{density}, \textit{pH}, \textit{sulphates}, \textit{alcohol}, and \textit{quality} as labels.

In the investigation of CNN adversaries, we used two different datasets called GENKI~\cite{GENKI-4K} and 11k Hands~\cite{afifi2016hands} (Figure~\ref{fig:dataset}).
GENKI is a dataset of $4000$ human faces that is labeled with \textit{head pose} and \textit{smile content.}
We replaced images that included multiple people or that were too low resolution for even humans to understand the image, and labeled \textit{gender} for each of the images.
We also normalized the sizes to $200 \times 200$ and converted the images to grayscale.

The main dataset is 11k Hands, a dataset of 11076 hands labeled with \textit{the subject ID}, \textit{gender}, \textit{age}, \textit{skin color}, \textit{left/right}, which we will call \textit{hand side}, \textit{dorsal/palmar}, \textit{accessories}, \textit{nail polish}, and \textit{irregularities}.
We only used the dorsal side ($5538$) images and combined the labels accessories and nail polish by an `or' operation to obtain a new label, \textit{decoration}.
Then, we reduced the size of the images to $200 \times 200$.
We also augmented the training data by flipping the images, changing to grayscale, adding noise, and shifting the images to help the network capture essential features.


\section{Results and Discussion}

\subsection{Linear Adversary Case}
\begin{figure*}[t!]
	\centering
  \begin{subfigure}{.32\textwidth}
    \centering
    \includegraphics[width=0.95\linewidth]{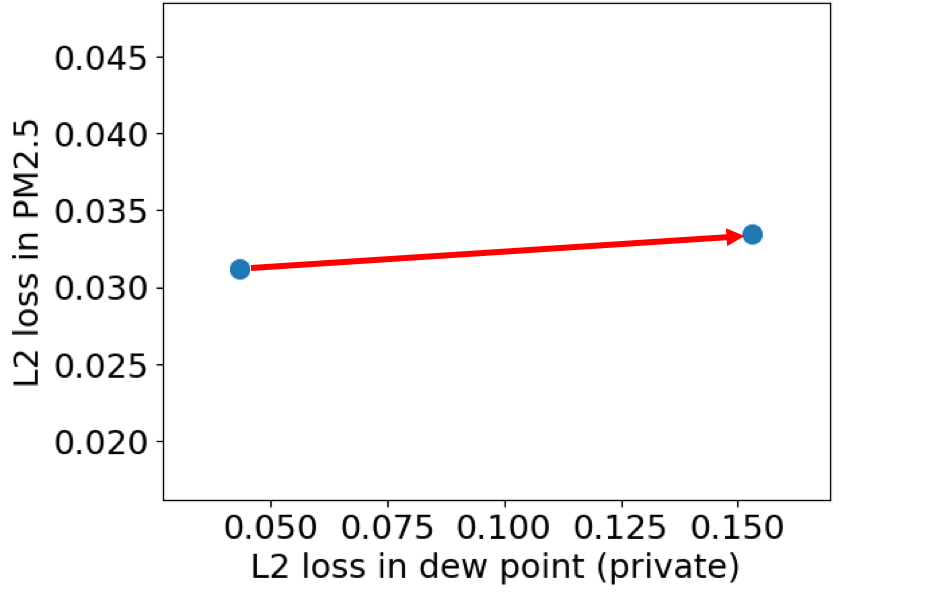}
    \caption{dew point (private) vs PM 2.5}
    \label{fig:beijing_priv}
   \end{subfigure}
  \begin{subfigure}{.32\textwidth}
    \centering
    \includegraphics[width=0.95\linewidth]{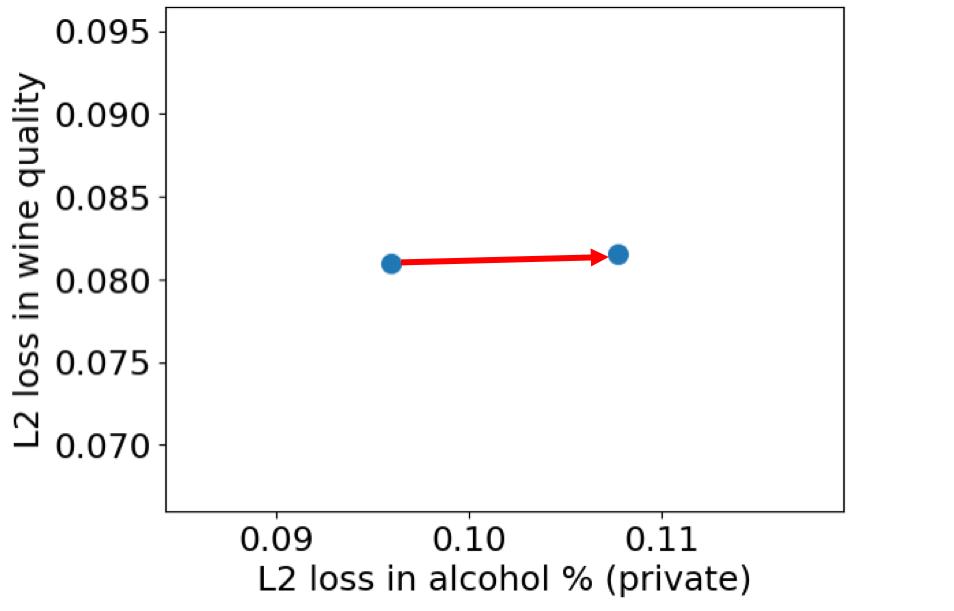}	
    \caption{alcohol \% (private) vs wine quality}
    \label{fig:wine_priv}
  \end{subfigure}
  \begin{subfigure}{.32\textwidth}
    \centering
    \includegraphics[width=0.95\linewidth]{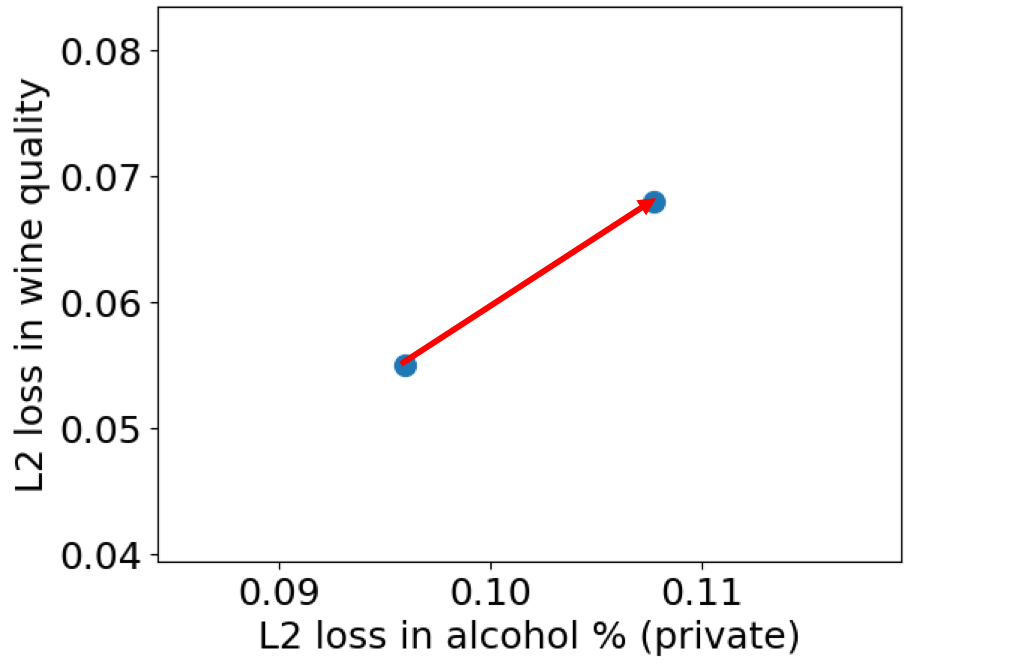}
    \caption{alcohol \% (private) vs pH}
    \label{fig:wine_priv2}
  \end{subfigure}
  
   \begin{subfigure}{.32\textwidth}
    \centering
    \includegraphics[width=0.95\linewidth]{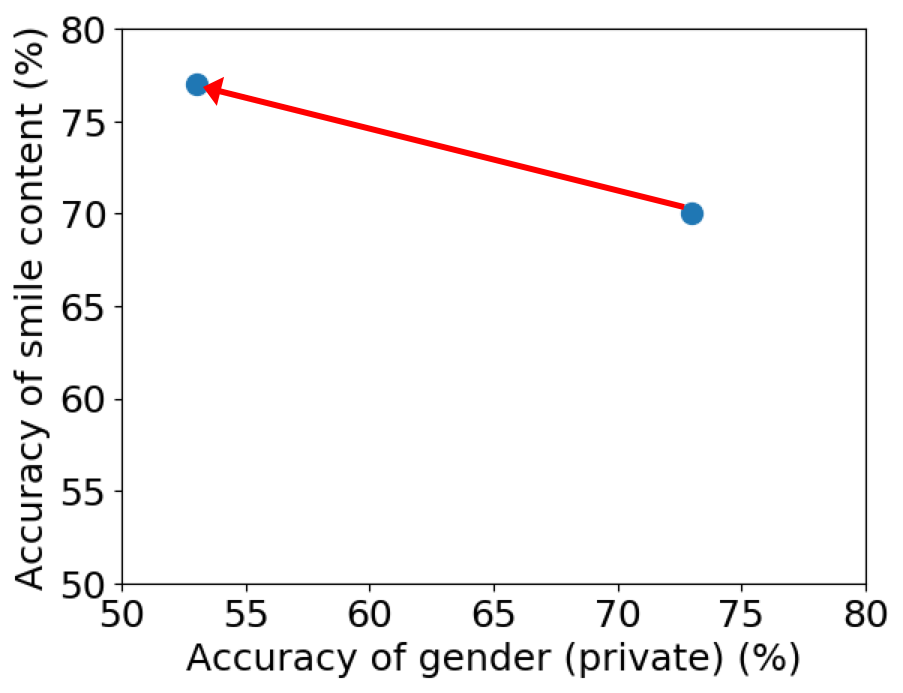}
    \caption{gender (private) vs smile content}
    \label{fig:genki_acc}
  \end{subfigure}
    \begin{subfigure}{.40\textwidth}
    \centering
    \includegraphics[width=0.95\linewidth]{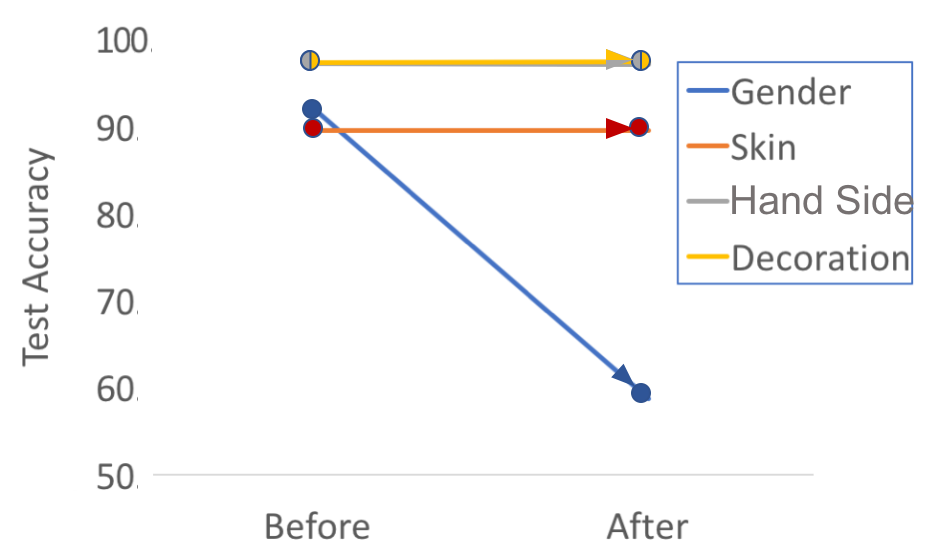}
    \caption{accuracies of labels for 11k Hands}
    \label{fig:11khand_acc}
  \end{subfigure}
\caption{Effect of privatization: (a) Beijing PM 2.5, (b, c) Wine (d) GENKI, and (e) 11k Hands }
\label{fig:greedy_priv}
\end{figure*}

\begin{figure}[ht]
	\centering
 	\includegraphics[width = \linewidth]{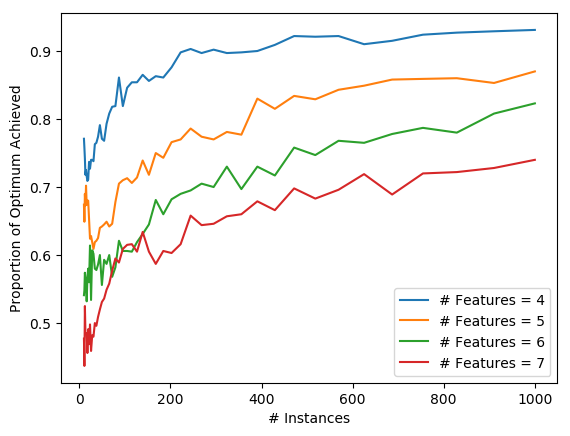}
	\caption{\label{fig:greedy_plot}Ratio of optimum achieved by greedy}
\end{figure}

Figure~\ref{fig:greedy_plot} shows the relationship between the number of instances and the proportion of runs in which the greedy approximation achieved the optimum for varying number of features, obtained from the randomly generated toy data.
The results indicate that the greedy approximation achieves the optimum more often as the number of instances increases.
Furthermore, the results seem to indicate that for greater number of features, we need more instances in order to accomplish the same proportion.

In the experiment with the Beijing PM 2.5 dataset, we set the dew point as the private label and PM 2.5 as the public label.
Figure~\ref{fig:beijing_priv} shows the change in the L2 loss for these two labels.
The loss of the dew point increased from $0.044$ to $0.153$, while the loss of PM 2.5 increased only from $0.031$ to $0.033$.
The features removed by greedy algorithm were in the order of: temperature, pressure, and day (of month).
These features are relevant to both labels, but whereas the dew point is directly related to the temperature and the pressure, PM 2.5 is only indirectly related to these features.

We set alcohol content as the private label and the rest of the labels as public labels in the experiment with the UCI Wine Quality.
Figure~\ref{fig:wine_priv} shows the change in the L2 loss for the private label, alcohol content, and a public label, wine quality.
The loss of the alcohol content increased from $0.096$ to $0.108$, while the loss of the wine quality increased only from $0.081$ to $0.082$.
The features removed by the greedy algorithm were in the order of: total $\text{SO}_{2}$, fixed acidity, and citric acid.
These features are highly relevant to the alcohol content.\footnote{$\text{SO}_{2}$ is a byproduct of fermentation, the cooler the origin of the wine, the higher the acidity and the lower the alcohol content.}

On the other hand, if we compare the loss of the private label, alcohol content, against that of another public label, pH, we experience a relatively high increase in the loss of the public label (Figure~\ref{fig:wine_priv2}).
Specifically, the loss of the pH increased from $0.055$ to $0.068$, which is comparable to the increase in the loss of the alcohol content. 
This is because two of the removed features, fixed acidity and citric acid, are highly correlated with both the pH and the alcohol content.
Here, we conclude that it is more difficult to preserve accuracy of the public labels that are dependent on the same features as the private label.

\subsection{CNN Adversary Case}
We used Google TensorFlow~\cite{45381} for handling neural networks.

For the GENKI dataset, we used two convolutional layers of sixteen $7 \times 7$ filters, each followed by a max-pool layers, and a fully connected layer of size $128$ for the adversary's network.
We used the same structure as the adversary's network for the privatizer's encoder, and used a fully connected layer of size $40000$ for the privatizer's decoder.

We set gender as the private label, and the smile content as the public label in this dataset.
The original test accuracy was 73\% for the gender and 70\% for the smile content. 
With the privatized images, the accuracy of the gender decreased to 53\%, while that of the smile content increased to 77\% (Figure~\ref{fig:genki_acc}).

For the 11k Hands dataset, we used three convolutional layers with two $3\times3$ filters, four $4\times4$ filters and eight $3\times3$ filters respectively, each followed by a max-pool layer, and finally a fully connected layer of size $32$ for the adversary's network. 
The pre-trained neural network for the protected labels has three convolutional layers with four $3\times3$ filters, eight $4\times4$ filters and sixteen $3\times3$ filters respectively, each followed by a max-pool layer, and finally a fully connected layer of size $64$. 
The encoder of the privatizer has the same structure as that of the protected labels, and decoder of the privatizer is symmetric to the encoder (Figure~\ref{fig:privatizer}).

\begin{figure}[t!]
	\centering
	\includegraphics[width = \linewidth]{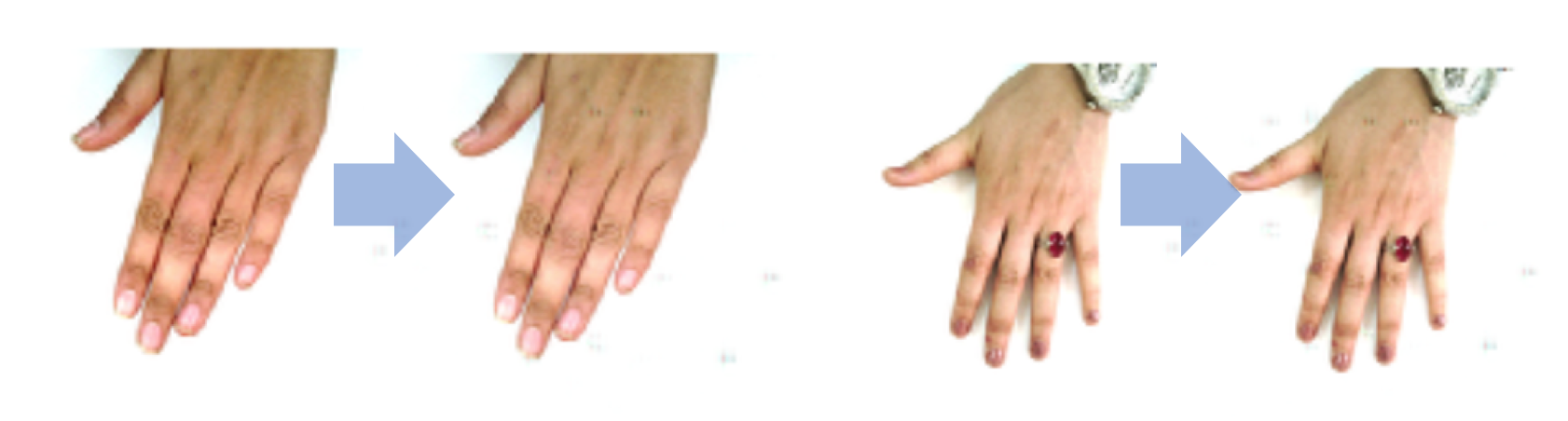}
	\caption{Original and privatized images}
	\label{fig:Noise}
\end{figure}

We set gender as the private label, decoration as the protected label, and the rest as public labels.
The accuracy achieved with the original data is 91.6\% (gender), 89.5\% (skin), 97.1\% (hand side), and 97.2\% (decoration).
With the privatized images, the accuracy for gender dropped to 58.7\%, while the accuracies for skin, hand side, and decoration changed to 89.5\%, 97.0\%, and 97.3\%, respectively (Figure~\ref{fig:11khand_acc}).
In sum, the accuracy for the private label dropped, whereas the accuracies for the other labels did not change significantly.
As we desired, the privatized images did not undergo notable distortion (Figure~\ref{fig:Noise}).

Theoretically the convergence occurs when the privatizer can cover all the private label-related features to a certain level. However, due to the limited computing resources and limited time, we were not able to attempt a thorough set of parameters with larger networks, and did not observe the convergence.

\section{Conclusion}
We present an efficient method of constructing a privatizer when the adversary is limited to linear models using a greedy approximation, which improves with the number of instances increasing.
Moreover, we found that the greedy algorithm removes features most correlated with the private label first, and that the algorithm preserves the predictability of public labels as long as they are sufficiently independent of the features that are relevant to the private label.

We also present a method for building a privatizer against a CNN adversary.
We were able to selectively lower the private label's accuracy while preserving other labels' accuracies against a fixed CNN adversary.
However, we lacked computational power for testing sufficiently many hyperparameters to achieve convergence.

\section{Future Work}
In the linear adversary case, theoretically bounding the performance of the greedy algorithm and finding how the relationship between features affects the optimum would solidify our findings.
Also, we only compressed the data with respect to the standard basis, but because any orthonormal basis can be transformed into a standard basis using an invertible matrix multiplication, we can easily generalize our results to any other orthonormal basis.
The natural next step is to find which orthonormal basis induces the best optimum.

Regarding the CNN adversary case, we will continue aiming for convergence of the algorithm by searching for hyperparameters in a larger space with greater computational power. In this case, the loss function should be calculated for all the history of the evolving adversary so that all the private attribute-related features are covered.
Afterwards, whether our method generalizes to more complicated neural networks remains a future work.


\section*{Acknowledgement}
We thank Peter Kairouz (kairouzp@stanford.edu), a postdoctorate scholar in the Department of Civil and Environmental Engineering, Stanford University, for his mentorship in this work.

\section*{Contributions}
Dae Hyun Kim has taken the lead on the linear adversary case and helped out with the CNN adversary case through general advice and help with finding parameters.
Taeyoung Kong has helped out with the linear adversary case through confirmation and scribing, and helped out with the CNN adversary case through general advice and help with finding parameters.
Seungbin Jeong has taken the lead on the CNN adversary case and helped out with the linear adversary case through general advice.

\bibliographystyle{ieeetr}
\bibliography{main}

\end{document}